\pdfoutput=1

\documentclass[11pt]{article}

\usepackage{ACL2023}

\usepackage{times}
\usepackage{latexsym}

\usepackage[T1]{fontenc}

\usepackage[utf8]{inputenc}

\usepackage{microtype}

\usepackage{inconsolata}

\usepackage{amsmath}
\usepackage{amssymb}
\usepackage{multirow}
\usepackage{algorithm}
\usepackage{graphicx}
\usepackage{algorithm}
\usepackage{courier}
\usepackage{booktabs}
\usepackage{multirow}

\DeclareMathOperator*{\argmax}{arg\,max}

\newcommand{\nop}[1]{}

\title{Towards Zero-Shot Frame Semantic Parsing with Task Agnostic Ontologies and Simple Labels}

\author{Danilo Ribeiro$^{1,2,}$\thanks{\ \ These authors contributed equally.}
~, Omid Abdar$^{1,*}$,
Jack Goetz$^1$, 
Mike Ross$^1$, 
Annie Dong$^1$,\\ 
\textbf{Kenneth Forbus}$^2$, 
\textbf{Ahmed Mohamed}$^1$ \\
$^1$Meta AI, \{jrgoetz, mikeross, asyd, ahmedkm\}@meta.com,\\
$^2$Northwestern University, \{dnribeiro, forbus\}@u.northwestern.edu,
}

\begin{document}
\maketitle
\begin{abstract}
Frame semantic parsing is an important component of task-oriented dialogue systems. Current models rely on a significant amount training data to successfully identify the intent and slots in the user's input utterance. This creates a significant barrier for adding new domains to virtual assistant capabilities, as creation of this data requires highly specialized NLP expertise. In this work we propose OpenFSP, a framework that allows for easy creation of new domains from a handful of simple labels that can be generated without specific NLP knowledge. Our approach relies on creating a small, but expressive, set of domain agnostic slot types that enables easy annotation of new domains. Given such annotation, a matching algorithm relying on sentence encoders predicts the intent and slots for domains defined by end-users. Extensive experiments on the \texttt{TopV2} dataset shows that our model outperforms strong baselines in this \textit{simple labels} setting.
\end{abstract}

\section{Introduction}

Frame semantic parsing is an important sub-problem with many applications, and in particular is critical for task-oriented assistants to identify the desired action (intent) and specific details (slots) \citep{coucke2018snips}. This is typically modeled as a semantic parsing problem (usually solved via a combination of ML and rules) with custom ontology that reflects capabilities of the system. Creation of this custom ontology, and annotation of consistent data is highly non-trivial, and typically requires specialized skills. This limits extension of the ontology and generation of parsing data to a small group of experts.

\begin{figure}[t!]
    \centering
    \includegraphics[scale=.45]{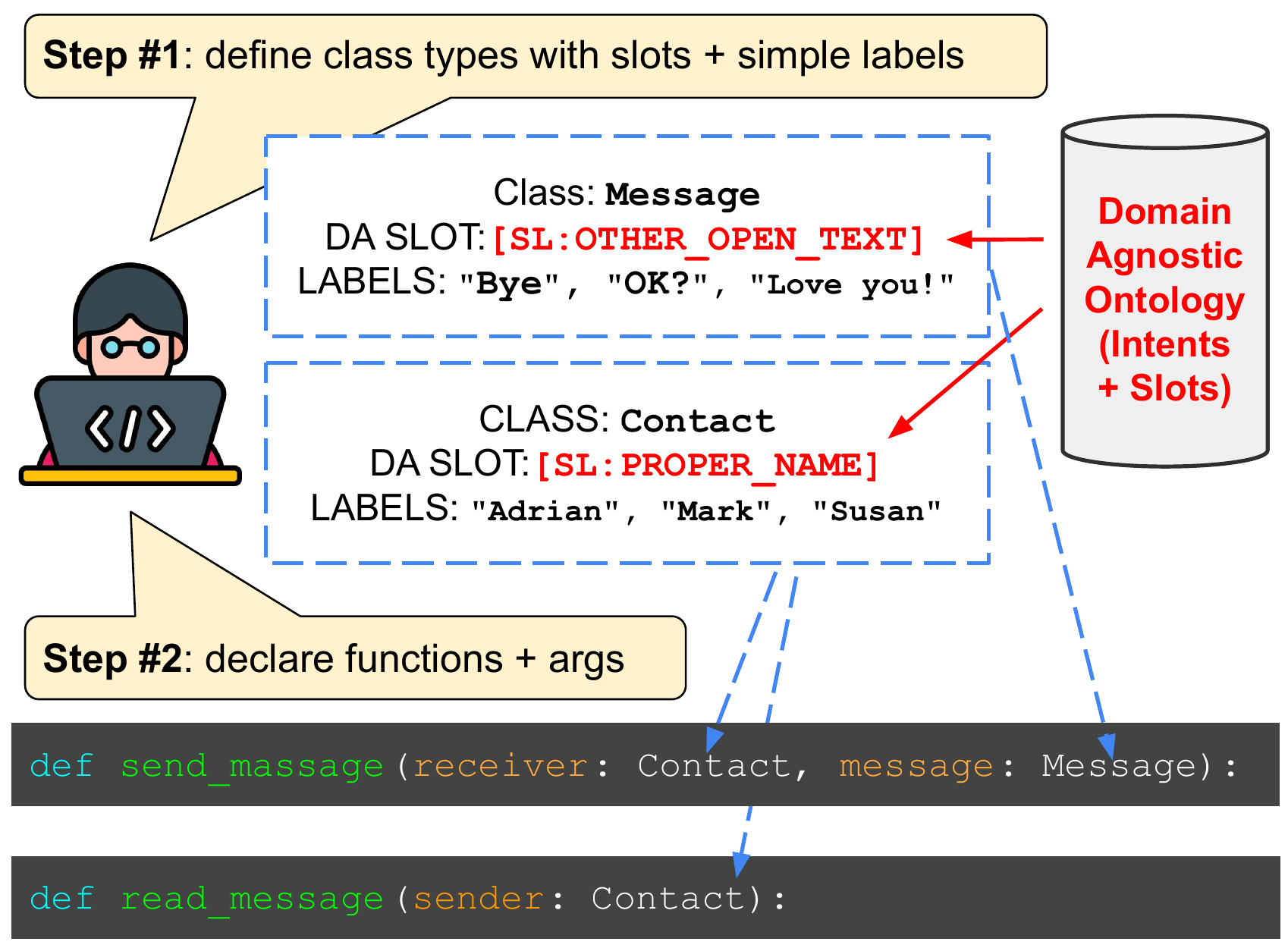}
    \caption{Illustration of proposed OpenFSP framework with a domain agnostic ontology and simple labels provided by the software developer (natural language textual examples). OpenFSP can facilitate the development of new domains and automatic construction of new ontologies by decoding functions or API specifications. }
    \label{fig:motivation}
\end{figure}

On the other hand, intent-slot concepts map well to functions and arguments of an API call, a paradigm well understood by software developers. Therefore, making extension to the parser is the primary blocker to enabling support for new capabilities (domains) within a task-oriented assistant system. As shown in Figure \ref{fig:motivation}, our goal is to enable non-NLP experts to define allowed intent-slot combinations, and provide a small amount of non-NLP specialized labels, which we call \textit{simple labels}. These data enables the creation of a parser for those intent-slot combinations. This new problem definition lies somewhere between zero-shot and few-shot learning. It requires zero fully annotated semantic parse examples, but does require some human produced labels.

To this end, we develop a framework called \textit{\textbf{Open} \textbf{F}rame \textbf{S}emantic \textbf{P}arser} (OpenFSP). OpenFSP takes as input the developer's defined functions and their annotations to augment an existing assistant system for new tasks. Underlying OpenFSP is a two module model consisting of a \textit{general semantic parser} and a \textit{matching module}. The general semantic parser can identify the intent and slots according to the pre-defined domain agnostic ontology, while the matching module will take this intermediate representation and match them to the specific function and arguments defined in the domain specific ontology.

In summary, our contributions are: (1) we formalize a new framework, namely OpenFSP, that allows for easy development of new domains without much expertise knowledge from software developers (2) we define a general-purpose domain agnostic ontology by analysing the semantic similarity of slots from \texttt{TopV2} \citep{chen-etal-2020-low}, a well established task-oriented semantic parsing dataset (3) we propose an approach consisting of a parser and a matching module that can outperform strong baselines in the \textit{simple labels} setting.

\section{Related Work}

\paragraph{Data-Efficient Semantic Parsing}{
In one of the first attempts to use data-efficient methods to perform frame semantic parsing, \citep{Bapna2017TowardsZF} applied recurrent neural networks to perform semi-supervised intent classification and slot filling (IC/SF) while leveraging natural language descriptions of slot labels. The work of \citet{krone-etal-2020-learning} more clearly formalized few-shot learning for IC/SF, while providing a few-shot split of three public datasets. \citet{wilson2019analogical} implemented and deployed a kiosk virtual assistant, which could handle multiple modalities by learning from a few examples using analogical learning.

Multiple low resource IC/SF approaches were proposed \citep{chen-etal-2020-low, desai2021low, yin-etal-2022-ingredients, basu-etal-2022-strategies}. All of these approaches either rely on a non-trivial amount of training data (hundreds to thousands of examples), or use a fixed set of intents and slots, making it harder to adapt to new domains. Our matching module shares some similarities with retrieval based systems \citep{yu-etal-2021-shot, shrivastava2022retrieve}, however these methods learn from standard utterance and semantic frames, instead simple textual labels for each slot and intent. This is an important differentiator, as the annotation of even a small number of semantic frames requires overcoming a significant knowledge barrier. 




}

\paragraph{Sentence Encoders for Language Understanding}{
One key aspect of our matching module is to encode the textual spans using sentence encoders. These models have the advantage of working well in low resource settings, and have been used in many applications including natural language inference \citep{conneau-etal-2017-supervised}, semantic textual similarity \citep{reimers-gurevych-2019-sentence}, dense passage retrieval \citep{karpukhin-etal-2020-dense, izacard-grave-2021-leveraging}, natural language explanations \citep{neves-ribeiro-etal-2022-entailment}, and many others. The idea of using sentence embeddings for text classification has been previously explored \citep{perone2018evaluation, piao2021scholarly}. Most notably, the work of \citet{tunstall2022efficient} proposed \textsc{SetFit}, a prompt-free model that can learn text classification tasks from a handful of examples. However, none of these works directly applied to semantic parsing which require multiple consistent predictions from the input utterance.
}

\paragraph{Task-Oriented Ontologies}{Task-oriented dialogue systems are natural language interfaces for a system that parses user utterances. A common approach in these systems is the use of ontologies to represent domain information \citep{wessel2019ontovpa}. In general, such ontologies can be created manually using rule-based techniques, which is a highly accurate but time consuming and expensive approach that usually requires domain experts \citep{meditskos2020converness}, or via ontology induction approaches using machine learning \citep{poon2010unsupervised}. While these approaches both involve trade-offs, curated ontologies can also be created via simplification of an existing ontology based on custom needs \citep{Kutiyanawala2018TowardsAS, laadidi_10.1145_3178461.3178483}. On the other hand, our work simplifies the ontological creation by abstracting the existing functions and arguments defined by the application itself.
}


\begin{figure*}[t!]
    \centering
    \includegraphics[scale=.63]{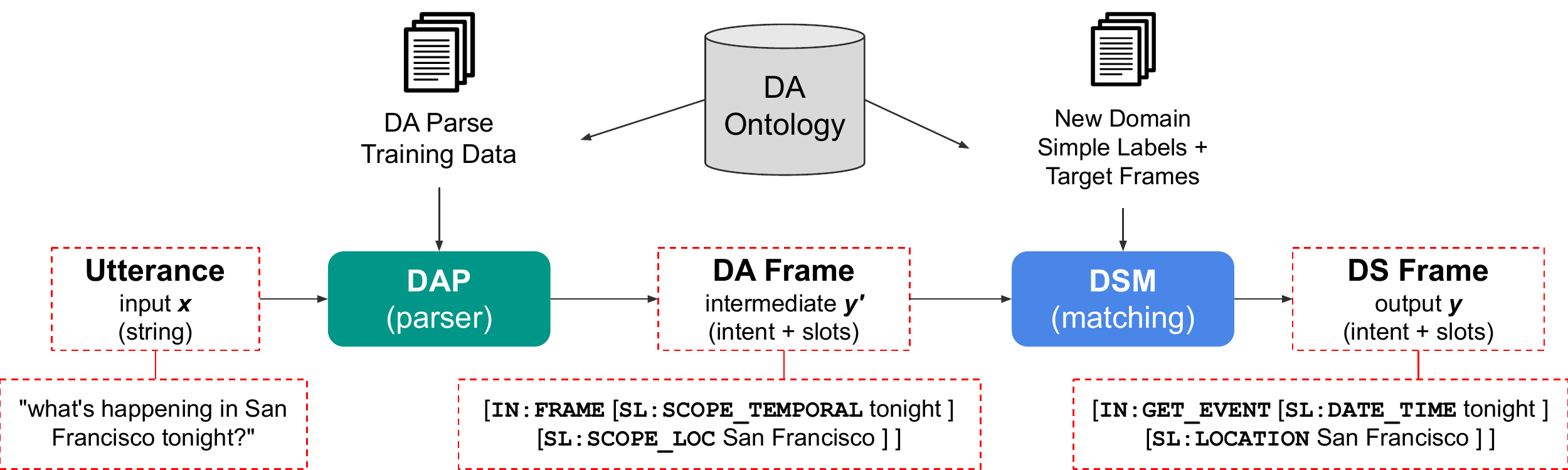}
    \caption{System overview with two components, namely the \textit{domain agnostic parser} (DAP) and the \textit{domain specific matching} (DSM).}
    \label{fig:system_overview}
\end{figure*}

\section{Problem Definition}

The standard frame semantic parsing has two main tasks which maps the input utterance $\boldsymbol{x}$ with tokens $x_{1}, \dots, x_{n}$ to some structured output frame $\boldsymbol{y}$. For the \textit{slot-filling} task, the output frame $F$ consists of a set of $m$ non-overlapping spans and their respective labels $F = \{(s_i, e_i, l_i)\}_{i=1}^{m}$ indicating that subsequence $\boldsymbol{x}_{[s_i:e_i]}$ has label $l_i \in \mathcal{L}$, where $\mathcal{L}$ is the set of possible labels (e.g., the text span ``noon tomorrow'' has the label \texttt{SL:DATE\_TIME}). The \textit{intent classification} assigns a label to the whole utterance. For simplicity, we assume that the intent can be thought of as another slot filling with $s_i = 0$ and $e_i = n$, with $l_i$ as the intent type.

To simplify annotation efforts, we define an ontology with a set of domain-agnostic labels $\mathcal{L}_A$, where $|\mathcal{L}_A| << |\mathcal{L}|$. These domain-agnostic labels can be interpreted as generic ``label types'', with an existing many-to-one mapping $\boldsymbol{\psi} : \mathcal{L} \rightarrow \mathcal{L}_A$ between the two sets. This mapping is further described in Appendix \ref{sec:app-domain-agnostic-ontology}. In our proposed simple label setting, training data for new domains will not include $\boldsymbol{x}$, but only a subsequence of the tokens defined by the frame spans. The number of examples for each slot will be relatively small, with 5 to 50 examples per slot type. The set of all possible domain specific target frames $F^{*}$ (i.e., defined functions and their arguments) is assumed to be known a priory.

\section{Approach}

Our model is comprised of two main components, the \textit{domain agnostic parser} (DAP) and the \textit{domain specific matching} (DSM). The system overview is shown in Figure \ref{fig:system_overview}. The DAP is trained to take the input utterance $x$ and output a domain agnostic frame $F_A$, where span labels belong to $\mathcal{L}_A$. Afterwards the DSM module will score the potential domain specific frames from $F^{*}$ according to their similarity to the domain agnostic frame $F_A$.

The DAP module can be any frame semantic parser. Ours is built on a span-pointer network \citep{shrivastava-etal-2021-span-pointer}, which is a non-autoregressive parser that replaces the decoding from text generation with span predictions. It is trained on domain agnostic data obtained from existing domain specific data using the slot label function $\boldsymbol{\psi}$.

The DSM module has to learn a similarity function between a text span and its slot label (same applies to intents). Since the slot label scoring function has to be done in a few-shot setting, the DSM module uses a sentence encoder $\phi$ \citep{reimers-gurevych-2019-sentence} with a classification head on top. The module modifies a pre-trained transformer language model fine tuned to output semantically meaningful sentence embeddings.

The sentence encoder $\phi$ is tailored to work on generic text, minimizing the distance between semantically similar sentences while maximizing the distance of dissimilar sentences. We use a classification head $H$ over the produced sentence embedding of a given text span $\boldsymbol{x}$. Therefore, the probability score of a label $l_i$ can be computed as follows:

\begin{equation}
  P(l_i | \boldsymbol{x}) = \frac{exp(H (\phi(\boldsymbol{x}))_i)}{\sum_{j=1}^{|\mathcal{L}|} exp(H( \phi(\boldsymbol{x}))_j) }
\label{eq:label-probability}
\end{equation}

Note that the input training data does not contain fully annotated domain specific frames (only text and their intent-slot labels). For this reason, the DSM module has to aggregate these individual intent and slot probability scores to predict the full frame scores. The similarity score for a target domain specific frame $F = \{(s_i, e_i, l_i)\}_{i=1}^{|F|}$ given a query domain agnostic parse $F_A = \{(s^A_i, e^A_i, l^A_i)\}_{i=1}^{|F_A|}$ is given by:


\begin{equation}
sim(F, F_A) =  \max_{F' \in \mathfrak{S}(F)} \frac{1}{|F'|} \sum_{i=1}^{|F'|} P(l_i | \boldsymbol{x}_{[s^A_i: e^A_i]})
\label{eq:similarity-score}
\end{equation}

Where $\mathfrak{S}(F)$ is the set of all slot permutations of $F$. To make the final prediction, we also check that the typing between domain agnostic and domain specific types match. Therefore DSM selects the best frame $F \in F^{*}$ using the formula:

\begin{equation}
\argmax_{F \in F^{*}} 
\begin{cases}
0 & \text{if } \exists i : \boldsymbol{\psi}(l_i) \neq l^{A}_{i} \\ 
sim(F, F_A) & \text{otherwise}
\end{cases}  
\label{eq:type-constraint}
\end{equation}

Note that for each new domain, only a very small set of parameters are trained (namely, the classification head $H$), which makes this approach computationally efficient.

\subsection{Data}



We conduct experiments using the \texttt{TopV2} \citep{chen-etal-2020-low}, a multi-domain dataset containing intent and slot annotations for task-oriented dialogue systems. We select a subset of the data points, ignoring the ones containing more than one intent (even though our method would also work for nested frames) or utterances with \texttt{IN:UNSUPPORTED} intent label. The final dataset contains a total of eight different domains (namely: \texttt{alarm}, \texttt{event}, \texttt{messaging}, \texttt{music}, \texttt{navigation}, \texttt{reminder}, \texttt{timer}, and \texttt{weather}). The dataset is then split into train (104278), evaluation (14509) and test (32654) sets. 


\section{Experiments}



\subsection{Evaluation Setup}

We conduct experiments to evaluate how well a model can adapt to a new unseen domain by leveraging the ``simple labels'' for this new domain. To simulate this adaptation process we perform multiple test rounds, where each of the eight \texttt{TopV2} domains are treated as unseen domains. The results for each left-out domain are averaged to obtain the final metrics. Each label $l_i \in \mathcal{L}$ of the unseen domain will be assigned a few textual examples that will be used for training. In our experiments we use 5, 10 and 50 examples per unseen domain's label. 


\subsection{Baselines}

The first two baselines are used as soft upper bounds. They do not follow the ``simple labels'' settings, and use the full training data for the new domains instead. The last two baselines are used as a away to evaluate the different implementation choices of the model, similar to ablation studies.

\paragraph{Majority Vote}{This simple model always selects the most commonly occurring intent and slots for a given domain. It uses the DAP module to generate $F_A$, and assigns the domain specific labels according to the number of occurrences of $F \in F^*$ (such that $|F| = |F_A|$) in the training data. 
}

\paragraph{Fully Supervised}{Uses the same architecture as the DAP (semantic parser) module, but it is trained using the full training data. Despite not being at all a fair comparison with the \textit{simple label} settings, we include these baseline results as a way to visualize the best-case scenario.
}

\paragraph{W.O. Head}{In this baseline we do not use a classification head $H$, instead we predict the class label by selecting the example with highest semantic similarity with the input text using the cosine similarity score.
}

\paragraph{W.O. Head + Type}{In this baseline we not only use the classification head, but also disregard the type constraint in Equation \ref{eq:type-constraint} such that the best frame is always $\argmax_{F \in F^{*}} \left( sim(F, F_A) \right)$.
}

\subsection{Implementation Details}
\label{sec:implementation-details}

The span-pointer architecture used by the DAP module is a encoder-decoder model based on RoBERTa \citep{Liu2019RoBERTaAR}, with the encoder containing 12 layers and the decoder containing 1 layer. We train the model for $85$ epochs using a learning rate of $1.67 \cdot 10^{-5}$. The sentence encoder used in the DSM module is built on top of a 36 layer XLM-R model \citep{conneau2019unsupervised} fine-tuned to capture general sentence similarity. When fine-tuning the models we used a machine containing two NVIDIA Tesla P100 graphics processing units. Note that the underlying models used are relatively small when compared to current large language models \citep{Sanh2019} and would be suitable for ``on-the-edge'' device computation.

\subsection{Results}

\begin{figure}[t!]
    \centering
    \includegraphics[scale=.56]{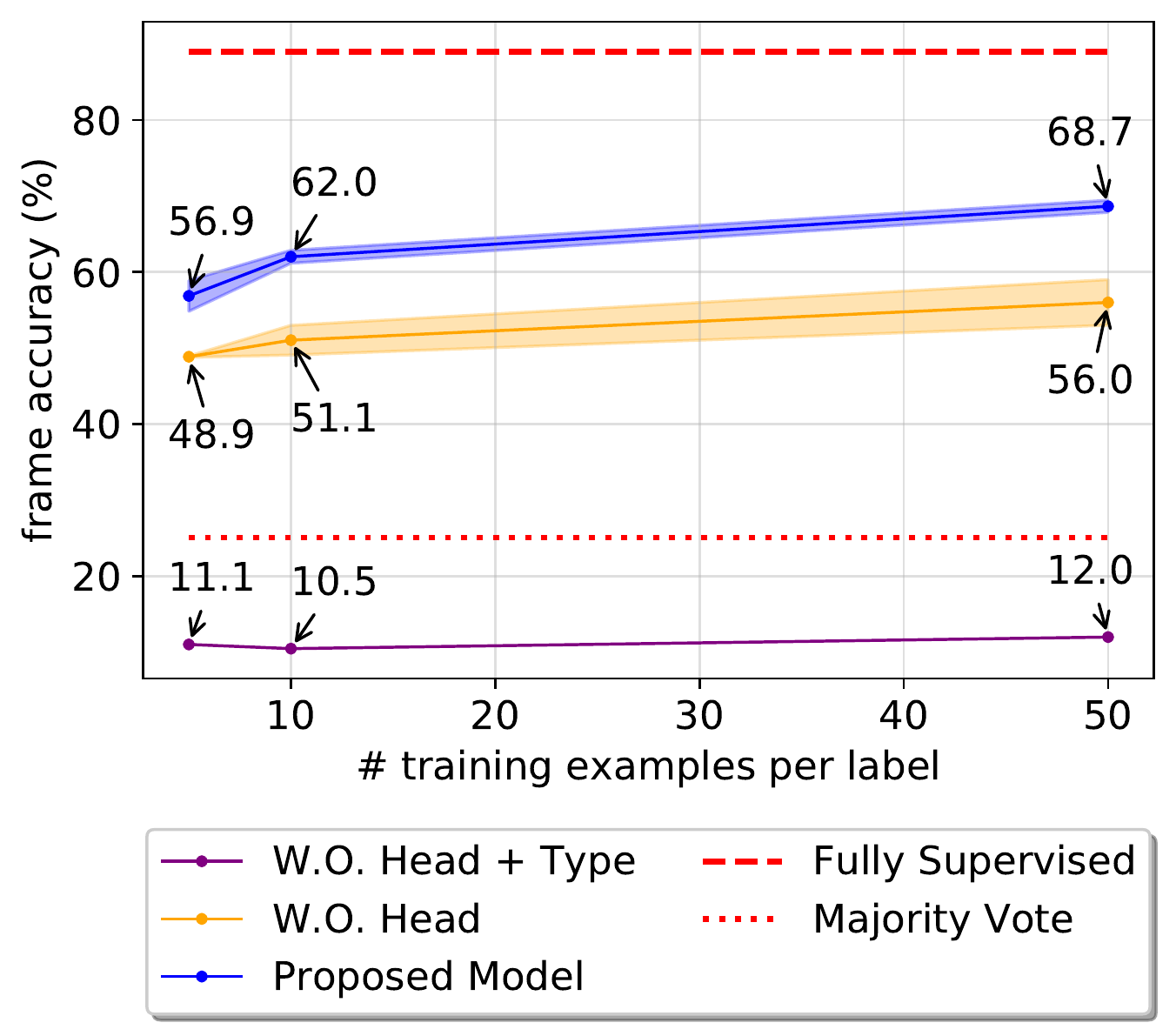}
    \caption{Main experiment results with different number of examples per label.}
    \label{fig:main_results}
\end{figure}

\begin{table*}[t!]
\centering
\setlength{\tabcolsep}{5pt}
\begin{tabular}{ l|cccccccc|c } 
\toprule
\textbf{Eval. Setting} & \textbf{messa.} & \textbf{alarm} & \textbf{music} & \textbf{event} & \textbf{navig.} & \textbf{remind.} & \textbf{timer} & \textbf{weath.} & \textbf{avg.} \\
\midrule
Standard & 86.7 & 67.6 & 55.5 & 73.9 & 60.6 & 58.5 & 72.7 & 79.6 & 69.4 \\
+    Golden Parse & 93.9 & 50.1 & 60.7 & 88.7 & 60.7 & 78.7 & 78.7 & 89.1 & 75.1 \\
+ Recall@3 & 92.9 & 78.8 & 65.0 & 82.2 & 75.0 & 68.9 & 78.9 & 90.2 & 79.0 \\
+ Intent Acc. & 96.9 & 82.3 & 76.6 & 97.9 & 78.6 & 61.5 & 80.4 & 85.7 & 82.5 \\
\bottomrule 
\end{tabular}
\caption{Break down of results by domain for different evaluation settings. The results shown correspond to a single run of our proposed model with 50 labels per example.}
\label{tab:result_breakdown}
\end{table*}

We performs multiple three test runs with different random seeds, influencing both the model initialization and the set of training examples per label. The main results are shown in Figure \ref{fig:main_results}. The baselines using the full training data (i.e., \textit{Fully Supervised} and \textit{Majority Vote}) are shown as dashed lines, and their results do not change according to the x-axis. The remaining results show the mean (and standard deviation as the shaded region) among the three test runs.

The results show that our proposed model outperforms most of the baselines, with results comparable to the \textit{fully supervised} baseline (77.9\% of its frame accuracy with 50 examples per label), even though it relies on a much smaller and simplified version of the training data. 

There are a few other takeaways. First, we notice that increasing the number of ``simple label'' training examples significantly improves the results, but results with only five training examples still yield decent results. More examples also seem to increases the variance of the models without a classification head. Second, the type filtering from Equation \ref{eq:type-constraint} is one key aspect of why the system can perform so well with so few example. Because of the generic nature of the domain agnostic ontology, filtering out invalid frames greatly reduces the size of the target space $F^*$, with a size reduction of 96\% for certain domains such as \texttt{messaging}.

\subsubsection{Results Break Down}

To obtain further insights on the model, we show the results broken down by domain in Table \ref{tab:result_breakdown}. We used our proposed model trained on 50 examples per label, and different evaluation settings described as follows. The \textit{Standard} settings are the same as the ones displayed in Figure \ref{fig:main_results}. The \textit{Golden Parse} assumes that the DAP module outputs only correct domain agnostic parses (i.e., the best-case scenario for the parser). The \textit{Recall@3} results uses the same model as the \textit{Standard} setting, but checks if the correct answer is in the top-3 scored matches (instead of top-1) from DSM. Finally, the \textit{Intent Acc.} setting evaluates if the model correctly predicts the intent of the given input utterance. These results help us answer the following questions:

\paragraph{How much error from the parser gets propagated?}{
We can notice from the \textit{Golden Parse} results that there is a modest improvement (8.2\% increase) in accuracy when using the gold test frames. This means that incorrect parses from the DAP module certainly propagates forward and improving the DAP module could certainly benefit the system as a whole.
}

\paragraph{Is the matching module nearly missing the right answer?}{
When looking \textit{Recall@3} results, we notice a significantly larger improvement in results (13.8\% increase). With an average mean reciprocal ranking among all domains of 74.9. Having a high frame score values in the top-3 is significant considering that on average the size among all domains for the target space $F^*$ is around 279.8 frames.
}

\subsection{Error Analysis and Future Work}

To understand the mistakes made by the OpenFSP system we perform some error analysis and suggest some possible improvement avenues for future work. For this analysis we 
use the development set and randomly sampled 100 output frames from different domains. We manually categorize the errors as follows. 

\paragraph{Parsing Errors}

We notice that 45\% of the analyses errors are due to parsing errors. This includes cases when the DAP module predicts an incorrect number of slots ($\sim$93\% among parsing errors) or when the number of slots are correct, but some of the slots have the incorrect labels ($\sim$7\% among parsing errors). A future direction could be to use the DAP and DSM modules to over-generate valid frames and rank \citep{varges2006overgeneration, zhang-wan-2022-mover}, which could circumvent parsing errors.

\paragraph{Intent Classification Error}

Another common error was the mislabeling of the utterance's intent, corresponding to 32\% of the manually categorized examples. This kind of error would often happen between semantically similar intents (e.g.,  \texttt{IN:PREVIOUS\_TRACK\_MUSIC} and \texttt{IN:REPLAY\_MUSIC} in the \texttt{music} domain) and with examples from \texttt{TopV2} that were labeled as unsupported  (e.g., \texttt{IN:UNSUPPORTED\_WEATHER} and \texttt{IN:UNSUPPORTED\_MUSIC}) that often have out of scope questions that are harder to classify (e.g., ``What is the hottest temperature this month''). One possible future direction would be to use Contrastive Learning \citep{chen2020simple, basu-etal-2022-strategies} that could improve the classification boundary of similar examples.


\paragraph{Slot Classification Error}

The last 23\% of the errors were due to slot type misclassification. Again, semantically similar slot types are more challenging to classify. For instance, in the \texttt{alarm} domain \texttt{SL:DATE\_TIME\_RECURRING}, \texttt{SL:DATE\_TIME}, \texttt{SL:PERIOD} and \texttt{SL:DURATION} were particularly hard to classify since they were all part of the same domain agnostic type \texttt{SL:SCOPE\_TEMPORAL}. Another common issue was identifying proper names ($\sim$21\% of the slot classification errors), including artist, event, album, and playlist names. A future direction would be to integrate a named entity recognition module to help classify slots involving proper names.

\section{Conclusion}

In this work we propose OpenFSP, a framework designed to simplify the process of adapting an existing task-oriented dialogue system to new domains. This framework enables non-experts to automatically build new domain ontologies from well defined software engineering concepts such as functions and arguments. We define a general-purpose domain agnostic ontology, that when combined with textual examples of new slots and intents (which we call \textit{simple labels}), provides sufficient data to adapt the system to a new domain.

Finally, we propose a two-module system that can use these simple labels to reasonably parse input utterances into the domain specific frames. Our experiments show that the proposed model outperforms strong baselines and is able to obtain results comparable with a fully supervised model (achieving 77.9\% of its semantic frame accuracy). We hope that our work will facilitate the development of new assistant capabilities, allowing end-users to interact with more software applications through natural language.

\section*{Limitations}
Domain-agnostic (DA) slots represent the overal semantic space covered by underlying TopV2 slots. Given their coarse-grain nature, DA slots are likely to be distributed more or less evenly across all domains. This assumption is key when training the parser on data that does not contain a particular target domain. In addition, we find that our approach is also sensitive to the types of linguistic structures accounted for by each DA slot and works best when these structures are consistent across domains.

More specifically, we conducted a series of leave-one-out (LOO) experiments where a separate parser was learned for each domain using training data from all other domains and then tested on test data from the domain in question exclusively. Error analysis of 100 randomly selected predictions in our LOO model for the “alarm” domain revealed that 32\% of the errors were utterances such as “I want alarms set for next Monday at 6.00am \underline{and} 7.00am” where the model predicted [\texttt{SL:SCOPE\_TEMPORAL} for next monday at 6.00am ] and [\texttt{SL:SCOPE\_TEMPORAL} 7.00am ] ] whereas a compound [\texttt{SL:SCOPE\_TEMPORAL} for next monday at 6.00am and 7.00am ] ] slot was expected. Upon closer inspection, we observed that these [\texttt{SL:SCOPE\_TEMPORAL} X and Y] compound nominal constructions appear predominantly (83.3\%) in the alarm domain, hence causing inaccurate predictions in a LOO model for this domain. 

For these types of errors to be mitigated in our approach, [\texttt{SL:SCOPE\_TEMPORAL} X and Y] constructions would need to either be more evenly distributed across all domains, or re-annotated as [ [\texttt{SL:SCOPE\_TEMPORAL} X] and [\texttt{SL:SCOPE\_TEMPORAL} Y] ] in the alarm domain to improve homogeneity in the data.

\section*{Ethics Statement}
No private data or non-public information was used in this work.


\bibliography{anthology,custom}
\bibliographystyle{acl_natbib}

\appendix

\section{Appendix}

\subsection{Domain Agnostic Ontology}
\label{sec:app-domain-agnostic-ontology}
To create our domain-agnostic ontology, we manually reviewed all existing slots in the \texttt{TopV2} ontology and categorized the semantic nature of each slot. We then conflated semantically-similar \texttt{TopV2} slots under overarching, domain-agnostic terms that cover the overall semantic space of the underlying \texttt{TopV2} slots. For instance, we categorize slots that indicate the user is requesting a "to-do item", "reminder," and "alarm" as roughly the overall "deliverable" item that is being requested, hence conflating the domain-specific slots \texttt{SL:TODO}, \texttt{SL:METHOD\_TIMER}, and \texttt{SL:ALARM\_NAME} under the domain-agnostic slot \texttt{SL:DELIVERABLE}. Table \ref{tab:domain_agnostic_ontology} contains the mapping between the \texttt{TopV2} ontology and the domain agnostic ontology used in this work.

\begin{table*}[t]
\centering
\small
\setlength{\tabcolsep}{4pt} 
\renewcommand{\arraystretch}{1.4} 
\begin{tabular}{|c|c|}

\hline
 
\textbf{Domain-agnostic (DA) slots} & \textbf{Domain-specific (DS) slots} \\

\hline

SL$:$DELIVERABLE & SL$:$TYPE\_REACTION, SL$:$TODO, SL$:$TODO\_NEW \\
& SL$:$METHOD\_TIMER, SL$:$TIMER\_NAME, SL$:$ALARM\_NAME  \\

\hline

SL$:$RECIPIENT & SL$:$RECIPIENT, SL$:$PERSON\_REMINDED\_ADDED, \\
& SL$:$PERSON\_REMINDED\_REMOVED, SL$:$PERSON\_REMINDED, \\
& SL$:$ATTENDEE\_REMOVED, SL$:$ATTENDEE\_ADDED \\

\hline

SL$:$SCOPE\_TEMPORAL & SL$:$DATE\_TIME, SL$:$DATE\_TIME\_RECURRING, \\
& SL$:$DURATION, SL$:$PERIOD, \\
& SL$:$RECURRING\_DATE\_TIME, SL$:$TIME\_ZONE, \\
& SL$:$DATE\_TIME\_DEPARTURE, SL$:$DATE\_TIME\_ARRIVAL, \\
& SL$:$FREQUENCY, SL$:$RECURRING\_DATE\_TIME\_NEW, \\
& SL$:$DATE\_TIME\_NEW, SL$:$SCOPE\_TEMPORAL\_RECURRING \\

\hline

SL$:$SCOPE\_LOC & SL$:$LOCATION, SL$:$POINT\_ON\_MAP, \\
& SL$:$LOCATION\_HOME, SL$:$LOCATION\_USER, \\
& SL$:$LOCATION\_MODIFIER, SL$:$WAYPOINT\_ADDED, \\
& SL$:$LOCATION\_WORK \\

\hline 

SL$:$SCOPE\_DISAM & SL$:$ORDINAL, SL$:$TYPE\_CONTENT, SL$:$GROUP, \\
& SL$:$RESOURCE, SL$:$CONTENT\_EMOJI, \\
& SL$:$TYPE\_CONTACT, SL$:$MUTUAL\_EMPLOYER, \\ 
& SL$:$MUTUAL\_SCHOOL, SL$:$TYPE\_INFO, \\
& SL$:$MUTUAL\_LOCATION, SL$:$CONTACT\_RELATED, \\
& SL$:$MUSIC\_GENRE, SL$:$UNIT\_DISTANCE, \\ 
& SL$:$WEATHER\_TEMPERATURE\_UNIT, SL$:$MEASUREMENT\_UNIT, \\ 
& SL$:$METHOD\_RETRIEVAL\_REMINDER \\ 
\hline

SL$:$OTHER\_OPEN\_TEXT & SL$:$CATEGORY\_EVENT, \\
& SL$:$SEARCH\_RADIUS, SL$:$ATTRIBUTE\_EVENT, \\ 
& SL$:$CATEGORY\_LOCATION, SL$:$NAME\_EVENT, \\ 
& SL$:$ATTENDEE, SL$:$ATTENDEE\_EVENT, \\ 
& SL$:$TYPE\_RELATION, SL$:$ORGANIZER\_EVENT, \\ 
& SL$:$TAG\_MESSAGE, SL$:$CONTENT\_EXACT, \\ 
& SL$:$MUSIC\_TYPE, SL$:$MUSIC\_TRACK\_TITLE, \\ 
& SL$:$MUSIC\_ALBUM\_TITLE, SL$:$MUSIC\_PLAYLIST\_TITLE, \\ 
& SL$:$MUSIC\_RADIO\_ID, SL$:$METHOD\_TRAVEL, \\ 
& SL$:$JOB, SL$:$WEATHER\_ATTRIBUTE, \\ 
& SL$:$OBSTRUCTION\_AVOID, SL$:$ROAD\_CONDITION\_AVOID, \\ 
& SL$:$ROAD\_CONDITION \\ 
\hline
        
SL$:$NUMS & SL$:$AMOUNT, SL$:$AGE \\
\hline

SL$:$PROPER\_NAME & SL$:$NAME\_EVENT, SL$:$CONTACT, \\ 
& SL$:$ORGANIZER\_EVENT, SL$:$SENDER, \\ 
& SL$:$MUSIC\_TRACK\_TITLE, SL$:$MUSIC\_PROVIDER\_NAME, \\ 
& SL$:$MUSIC\_ALBUM\_TITLE, SL$:$MUSIC\_ARTIST\_NAME, \\ 
& SL$:$SOURCE, SL$:$DESTINATION, SL$:$PATH, SL$:$PATH\_AVOID, \\ 
& SL$:$WAYPOINT\_AVOID, SL$:$LOCATION\_CURRENT, \\ 
& SL$:$PATH\_AVOID, SL$:$WAYPOINT\_AVOID, \\ 
& SL$:$LOCATION\_CURRENT, SL$:$WAYPOINT, \\
& SL$:$ATTENDEE, SL$:$NAME\_APP \\ 
\hline

\end{tabular}

\caption{Mapping between \texttt{TopV2} ontology and our proposed domain agnostic ontology.}
\label{tab:domain_agnostic_ontology}
\end{table*}



\end{document}